\pdfoutput=1
\documentclass{article}

\usepackage[preprint]{arxiv_preprint}

\usepackage[utf8]{inputenc}
\usepackage[T1]{fontenc}
\usepackage[hypertexnames=false]{hyperref}
\usepackage{url}
\usepackage{booktabs}
\usepackage{amsfonts}
\usepackage{nicefrac}
\usepackage{microtype}
\usepackage{xcolor}
\usepackage{amsmath}
\usepackage{amssymb}
\usepackage{graphicx}
\usepackage{multirow}
\usepackage{tikz}
\usepackage{float}
\usepackage{capt-of}

\title{Efficient Spatio-Temporal Grounding with Multimodal Large Models via Second-Level Tracking and RL Verification}

\author{%
  Tianshu Zhang\thanks{Work done during internship at Z.ai.} \\
  Tsinghua University
  \And
  Yan Wang \\
  Tsinghua University
  \And
  Ji Qi\thanks{Corresponding authors.} \\
  Tsinghua University
  \And
  Lijie Wen\footnotemark[2] \\
  Tsinghua University
}

\begin{document}

\maketitle

\begin{abstract}
Spatio-temporal grounding in long videos requires precise temporal localization and robust object tracking conditioned on natural-language queries. While recent vision-language models (VLMs) show strong reasoning ability, directly applying frame-by-frame inference to long sequences is computationally expensive and unstable. We propose a practical pipeline that shifts from frame-level to second-level tracking and performs cross-second smoothing to preserve continuity while reducing sequence length. To improve reasoning supervision, we synthesize chain-of-thought style trajectories using advanced multimodal models for temporal localization and target selection, and replace generated spatio-temporal coordinates with ground-truth annotations to avoid noisy supervision. We further optimize the policy with reinforcement learning using a verifier based on $t\_\mathrm{IoU}+mv\_\mathrm{IoU}$. Experiments across multiple FPS settings show that our method achieves a strong trade-off between efficiency and localization quality.

\end{abstract}

\newcommand{\qwen}{Qwen3.5}
\newcommand{\seed}{Seed2.0-pro}
\newcommand{\kimi}{Kimi-k2.5}
\newcommand{\geminiflash}{Gemini3-Flash}
\newcommand{\geminipro}{Gemini3-Pro}

\section{Introduction}
Spatio-temporal video grounding (STVG) asks a model to localize a natural-language query in both time and space \citep{Su_2021_ICCV}. Given a video and a sentence such as ``the person in red picks up the bag'', the model must identify the event interval and track the referred target throughout it. The task combines temporal moment localization, referring expression comprehension, object tracking, and video-language reasoning, making it a natural benchmark for language-controllable video understanding.

STVG provides a more complete grounding interface than temporal localization, frame-level referring localization, or tracking alone. Temporal-only methods cannot identify where the described target is, while tracking methods do not solve language-conditioned search by themselves. STVG combines these requirements into a single task, making it useful for video retrieval, interactive annotation, human activity understanding, robot perception, and video agents. Existing systems make progress by coupling proposal generation, temporal boundary prediction, detection, tracking, cross-modal fusion, and target-aware reasoning \citep{zhang2020objectawaremultibranchrelationnetworks,gu2025knowingtargettargetawaretransformer}, but many still rely on dense frame processing or specialized modules whose cost grows quickly with video length.

% Qiji: Introduction这里还一个图，展示已有模型的问题（逐帧追踪token爆炸、CoT没逻辑） vs 我们模型的方法（逐秒追踪+平滑、先时再空的CoT）
% 左边的已有模型大概是：一个多模态大模型（ViT+LLM），下方是输入的一堆视频帧 和 用户prompt（比如询问视频中粉色海豚的追踪结果），上方是模型的逐帧输出（一大堆坐标框，用数量展示token特别多，输出没有逻辑）。 右边的我们的模型大概是：同样的多模态大模型，下方是一堆视频帧 和 逐秒追踪的prompt（询问粉色海豚的逐秒追踪结果），上方是逐秒追踪的输出（不太多的坐标框，且输出有逻辑系，先时间上寻找，再空间上寻找，最后输出坐标框），上方逐秒追踪的输出上面再加一层平滑层，将没秒的结果反向分发给一秒内每一帧上
\begin{figure}[t]
    \centering
    \includegraphics[width=1.0\linewidth]{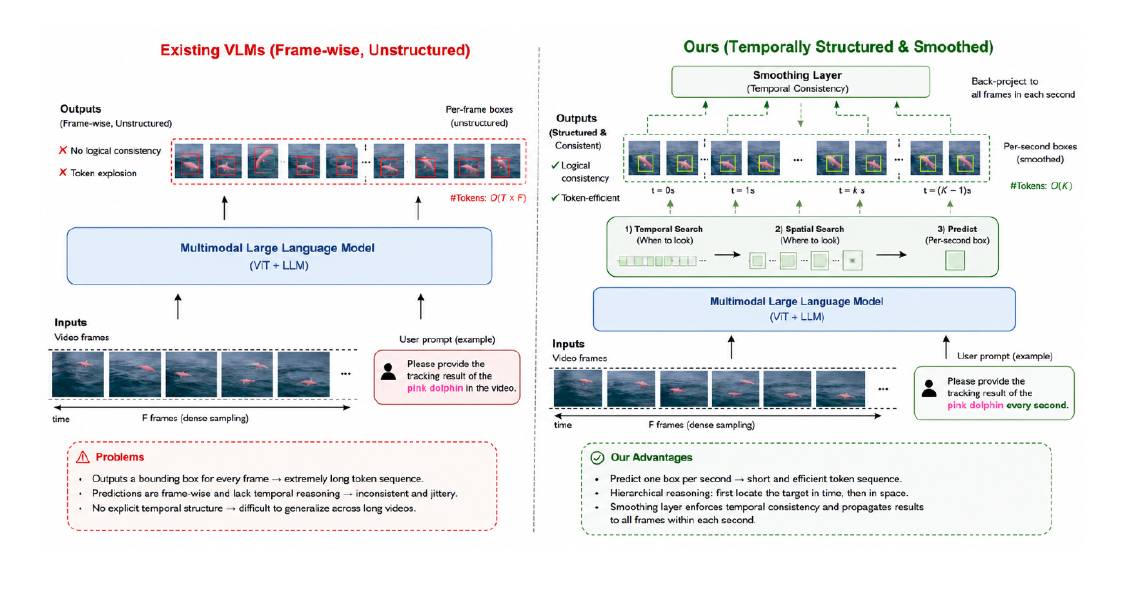}
    \caption{Comparison between our method and traditional VLMs. Left: existing VLMs performing per-frame tracking lead to token explosion. Right: our proposed per-second tracking mode with a post-smoothing strategy and the spatio-temporal reasoning paradigm.}
    \label{fig:intro_fig_example}
\end{figure}

The difficulty becomes more pronounced in long videos. A short query may describe a brief action inside long background context, and the referred object may remain visible before and after the event with different relevance. Dense sampling preserves boundary cues but increases visual tokens and tracking states; sparse sampling is cheaper but may miss decisive transitions. Even after the correct segment is found, appearance change, occlusion, fast motion, and distractors can cause identity switches or jitter.

Recent MLLMs offer a promising alternative because they provide stronger language understanding, broader visual knowledge, and structured intermediate reasoning. For STVG, this is appealing: complex queries often require parsing attributes, relations, actions, and event order before deciding when and where the target appears. MLLMs can connect such linguistic constraints to visual evidence and express decisions in a flexible output format.

However, there is still a clear gap between this promise and a robust STVG system. Direct frame-level MLLM grounding requires long visual contexts, fine temporal reasoning, and precise text-generated trajectories. One workaround is to input frames iteratively and track step by step \citep{fu2025omniptunleashingpotentiallarge}, but this turns long-video grounding into many sequential MLLM calls and makes the wall-clock cost too high. MLLM-based STVG therefore needs a formulation that preserves semantic reasoning while controlling sequence length and grounding noise.

Our work focuses on this practical missing piece: how should an MLLM-based STVG system handle long videos while producing usable temporal and spatial grounding? We argue that the basic reasoning unit should be the second rather than the frame. Many STVG queries describe actions and interactions that persist across short temporal neighborhoods; second-level units therefore shorten the sequence, preserve semantic event structure, and yield compact trajectories. A lightweight inter-second smoothing step restores continuity and reduces jitter.

The second-level formulation also changes the training problem. We use three stages: CPT for broad spatio-temporal grounding, supervised fine-tuning for structured target and event reasoning, and reinforcement learning for metric-aligned localization. During SFT, generated traces expose temporal and identity decisions, but final coordinates and segments are replaced with annotations; RL then scores complete predictions with temporal and motion-aware spatial overlap, aligning optimization with evaluation more directly than token likelihood alone.

Taken together, our formulation couples second-level tracking, ground-truth-corrected reasoning supervision, and verifier-driven RL. These components address long visual inputs, drift-prone traces, text-token coordinates, and metric-misaligned supervision. We evaluate on VidSTG subsets and HC-STVG, with FPS sensitivity and ablation studies analyzing the efficiency-quality trade-off.

Our contributions are summarized as follows:
\begin{itemize}
\item We formulate MLLM-based STVG around second-level tracking, reducing long-video sequence cost while maintaining track continuity through inter-second smoothing.
\item We develop a three-stage training pipeline that combines CPT, corrected reasoning SFT, and verifier-driven RL for spatio-temporal localization.
\item We construct reasoning supervision with generated decision traces and ground-truth coordinate replacement, separating useful semantic reasoning from unreliable generated geometry.
\item We optimize with a verifier based on $t_{\mathrm{IoU}}+mv_{\mathrm{IoU}}$ and evaluate across VidSTG subsets, HC-STVG, multiple FPS settings, and targeted ablations.
\end{itemize}

\section{Related Work}
\paragraph{Video grounding and spatio-temporal localization.}
Spatio-temporal video grounding (STVG) localizes both the temporal interval and frame-level trajectory of a target described by language. Early non-MLLM studies approached this through language-guided video retrieval, object referring, and sentence-guided video segmentation \cite{yamaguchi2017spatiotemporalpersonretrievalnatural, vasudevan2018objectreferringvideoslanguage, gavrilyuk2018actoractionvideosegmentation}. Later conventional methods rely on task-specific graph reasoning, two-stage localization, visual-linguistic Transformers, DETR-style tube prediction, consistency modeling, context-guided decoding, and open-vocabulary grounding \cite{tan2022augmented2dtantwostageapproach, yang2022tubedetrspatiotemporalvideogrounding, jin2022embracingconsistencyonestageapproach, lin2022stvgformerspatiotemporalvideogrounding, gu2024contextguidedspatiotemporalvideogrounding, wasim2024videogroundingdinoopenvocabularyspatiotemporalvideo}. These systems provide strong task-specific baselines, but they remain specialized cross-modal matching or tube regression pipelines with limited high-level reasoning and open-world knowledge.

\paragraph{Grounding in Multimodal Large Language Models.}
Recent MLLMs show promising grounding ability, but most work studies either spatial grounding in images or temporal grounding in videos. Spatial grounding models such as Qwen-VL \citep{bai2023qwenvlversatilevisionlanguagemodel} connect language to regions through location tokens, natural-language coordinates, image-caption-box alignment, referring mechanisms, or masks, while video MLLMs such as TimeChat \citep{ren2024timechattimesensitivemultimodallarge} and query-enrichment methods \citep{pramanick2025enrichdetectvideotemporal} localize relevant moments from language. Later MLLMs \citep{vteam2026glm45vglm41vthinkingversatilemultimodal,bai2025qwen3vltechnicalreport,vteam2026glm5vturbonativefoundationmodel} increasingly treat spatial and temporal grounding as core capabilities, yet they are still often evaluated separately \citep{cheng2025vstarbenchmarkingvideollmsvideo,meng2026videozerobenchprobinglimitsvideo}; methods that jointly predict temporal boundaries and spatial trajectories remain less common \citep{viditeam2026vidi25largemultimodalmodels,li2025videochatr1enhancingspatiotemporalperception,clark2026molmo2openweightsdata}. Recent systems such as LLaVA-ST \citep{li2025llavastmultimodallargelanguage} and SpaceVLLM \citep{wang2025spacevllmendowingmultimodallarge} explore fine-grained spatial-temporal outputs, and other work studies zero-shot STVG or reinforcement fine-tuning \citep{fu2025omniptunleashingpotentiallarge,meng2026openo3videogroundedvideoreasoning,yang2025unleashingpotentialmultimodalllms,gu2025thinkingboundingboxesenhancing}. However, many approaches depend on architecture changes \citep{chamiti2025refergptzeroshotreferringmultiobject,yang2026tracevisiontrajectoryawarevisionlanguagemodel,wang2024elysiumexploringobjectlevelperception} or external-tool pipelines \citep{gao202511,wu2026massmotionawarespatialtemporalgrounding,munasinghe2025videoglammlargemultimodalmodel,wu2026massmotionawarespatialtemporalgrounding}, limiting generality \citep{yang2025multiobjecttrackingretrievalllavavideo}. Our work instead studies joint spatial-temporal localization and RL enhancement without changing architecture or inference pipelines.

\section{Method}

\subsection{Problem Definition}
We study spatio-temporal video grounding (STVG) under a language-conditioned setting. Given an input video $V=\{f_1,\dots,f_T\}$ with frame rate $F$ and a natural-language query $q$, the model needs to output both the temporal interval in which the described event occurs and the spatial location of the referred target throughout that interval. The ground-truth annotation can be written as
\begin{equation}
y^{\ast}=(t_s^{\ast},t_e^{\ast},\mathcal{B}^{\ast}), \qquad
\mathcal{B}^{\ast}=\{b_k^{\ast}\}_{k=t_s^{\ast}}^{t_e^{\ast}},
\end{equation}
where $[t_s^{\ast},t_e^{\ast}]$ denotes the target temporal span and $b_k^{\ast}$ is the target box at time index $k$. In frame-level STVG, $k$ usually corresponds to a frame. This is precise, but the sequence length grows linearly with the number of frames and quickly becomes impractical for MLLMs on long videos.

We instead convert the video into second-level units $\mathcal{S}=\{S_1,\dots,S_K\}$, where $K=\lceil T/F\rceil$ and $S_k$ contains the frames falling into the $k$-th second. The model predicts
\begin{equation}
\hat{y} = (\hat{t}_s,\hat{t}_e,\hat{\mathcal{B}}), \qquad
\hat{\mathcal{B}} = \{\hat{b}_k\}_{k=\hat{t}_s}^{\hat{t}_e}.
\end{equation}
Here $[\hat{t}_s,\hat{t}_e]$ is a second-level temporal span and $\hat{b}_k$ is the predicted target state for second $k$. When original annotations are available at frame granularity, we convert them to second-level supervision by aggregating visible target boxes within each second. This representation keeps the event structure needed by STVG while reducing the number of visual units from $T$ frames to roughly $T/F$ seconds. It also matches the fact that many language queries describe actions, interactions, and state changes that persist across short temporal neighborhoods rather than isolated frames.

\subsection{Why Direct MLLM Grounding is Difficult}
MLLMs are attractive for STVG because they can parse complex referring expressions and reason over event context, but direct STVG asks them to solve continuous localization through long discrete generation. A straightforward design would feed many frames to the MLLM and ask it to generate a temporal span together with dense coordinates. This creates several mismatches between what STVG requires and what MLLMs are naturally optimized to produce.

Frame-level visual input quickly creates an expensive context problem: dense sampling preserves boundary cues but increases visual-token cost, while sparse sampling may remove the evidence that distinguishes the queried event from surrounding context. Direct grounding also lengthens the output, so generated reasoning and coordinate sequences can drift across similar frames, visible distractors, nearby temporal units, or language priors that are insufficiently grounded in visual evidence \citep{chen2024ictimageobjectcrossleveltrusted}. Finally, coordinates are a brittle text-generation interface for dense geometry: small numerical errors can produce jitter, and independently generated boxes do not guarantee stable identity. Since SFT optimizes token likelihood rather than temporal overlap, spatial overlap, or trajectory consistency, we use compact second-level outputs, corrected geometric supervision, and task-level RL rewards.

\begin{figure}[t]
    \centering
    \includegraphics[width=1.0\linewidth]{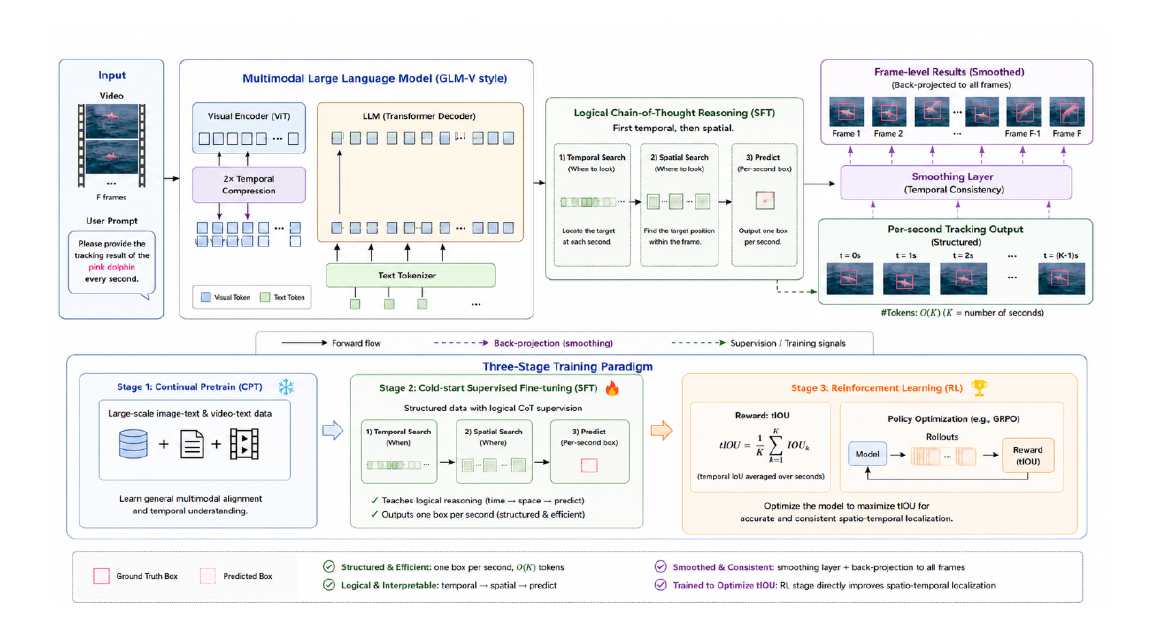}
    \caption{Model architecture and three-stage training. We train a GLM-V-style multimodal architecture to output per-second tracking results. The reasoning adopts a ``time-then-space" logical CoT. Training includes CPT, logical CoT-focused SFT, and RL for spatio-temporal grounding.}
    \label{fig:secotrack_method}
\end{figure}

\subsection{Three-Stage Training Pipeline}
Our method uses a three-stage pipeline designed around these failure modes and is built on a GLM-V-style architecture. CPT builds the basic ability to connect language, temporal units, and object locations. SFT teaches the model how to reason about the queried target and approximate temporal window. RL then directly optimizes numerical grounding quality with task-level rewards to fully unlock models' capablities. The stages are complementary: CPT provides broad visual grounding ability, SFT shapes the model's decision process, and RL aligns the final output with STVG metrics.

\subsection{CPT: Broad Spatio-Temporal Grounding}
\paragraph{Data construction.}
The CPT stage uses a mixture of STVG-style data, large-scale auxiliary tracking and segmentation data, and general video understanding data. In addition to video grounding annotations, we include a large amount of multi-object tracking (MOT) \citep{cui2023sportsmotlargemultiobjecttracking}, single-object tracking (SOT) \citep{müller2018trackingnetlargescaledatasetbenchmark,Huang_2021,fan2020lasothighqualitylargescalesingle}, and segmentation data \citep{miao2021vspw,ravi2024sam2segmentimages}. MOT data teaches the model to distinguish multiple instances, preserve identity under interactions, and reason about distractors. SOT data provides dense examples of following one target over time, which is important for stable trajectories after the temporal span has been selected. Segmentation data contributes fine-grained objectness and boundary information; we convert masks to boxes when training the box-output interface, while retaining the benefit of precise foreground supervision during representation learning.

Most videos in CPT are sampled at 1 FPS, which matches our second-level formulation and keeps the visual context affordable. We also mix in a small portion of higher-FPS videos to preserve distributional coverage and ensure that the model still has sufficient processing ability when evaluation videos contain denser temporal evidence. This mixture avoids over-specializing the model to a single 1-FPS regime.

All grounding-related data sources are converted into a unified second-level format. For tracking data, each training instance is formed by pairing a target description or template-derived prompt with a second-level trajectory. For segmentation data, prompts are constructed around object categories or referred regions, and the target mask or box is assigned to the corresponding second-level unit. For STVG data, the temporal span and trajectory are already aligned with language queries and are converted directly. Mixing these sources is important because STVG datasets alone are usually not large enough to teach robust tracking, object identity preservation, and spatial precision across diverse scenes at the CPT stage. We further add general video captioning and video question-answering data so that CPT does not collapse into a narrow coordinate-prediction task. These general video samples preserve the model's basic conversational ability and broad video understanding ability, which are necessary for interpreting open-ended user queries before producing grounded outputs.

\subsection{SFT: Reasoning Without Coordinate Drift}
\paragraph{Training data.}
The SFT stage uses $20$K examples sampled from the Vidstg and HC-STVG training sets. For each example, we use Gemini to generate a chain-of-thought style decision trace. The trace focuses on two questions: what object or person should be tracked, and what rough temporal window contains the described event. This supervision is intended to teach the model to make its grounding decision through explicit visual-language reasoning rather than by directly emitting coordinates from shallow pattern matching.

\paragraph{Reasoning format.}
We deliberately do not ask the model to output spatial coordinates inside the reasoning chain. This design keeps the trace focused on semantic decisions rather than dense geometry: the model describes the target, discriminative attributes, relevant actions, and approximate temporal evidence, while the final structured answer compactly provides the temporal span and trajectory supervised by ground truth. We also encourage the trace to spend more reasoning on locally ambiguous moments, such as adjacent temporal units with distractor objects or similar actions, instead of repeatedly verbalizing routine frame-level observations.

\paragraph{Coordinate correction.}
Although Gemini-generated traces are useful for semantic reasoning, we do not trust generated geometry as supervision. For any final structured output, temporal spans and boxes are replaced with ground-truth annotations. This produces a corrected sequence $\tau$ that preserves high-level reasoning while ensuring that all numerical labels are geometrically valid. The SFT loss is the standard next-token objective:
\begin{equation}
\mathcal{L}_{\text{sft}} =
-\mathbb{E}_{(\mathcal{S},q,\tau)}
\sum_u \log \pi_{\theta}(\tau_u\mid\tau_{<u},\mathcal{S},q).
\end{equation}
This stage teaches the expected response structure and reasoning behavior for STVG, while leaving metric-level localization quality to the RL stage.

\subsection{RL: Direct Optimization of Grounding Quality}

\paragraph{Reward design.}
For each prompt, the policy samples a group of candidate outputs. We parse each output into a temporal span and second-level trajectory, then score it with a verifier:
\begin{equation}
R = w_t\, t_{\mathrm{IoU}} + w_m\, mv_{\mathrm{IoU}}.
\end{equation}
The temporal reward $t_{\mathrm{IoU}}$ measures overlap between the predicted and ground-truth time windows:
\begin{equation}
t_{\mathrm{IoU}} =
\frac{|[\hat{t}_s,\hat{t}_e]\cap[t_s^{\ast},t_e^{\ast}]|}
{|[\hat{t}_s,\hat{t}_e]\cup[t_s^{\ast},t_e^{\ast}]|}.
\end{equation}
This term provides a direct learning signal for event localization before spatial trajectory quality is evaluated.

The motion-aware spatial reward $mv_{\mathrm{IoU}}$ evaluates whether the predicted trajectory follows the correct target during the relevant time span. We compute spatial overlap over valid second-level states and aggregate it across the matched temporal region:
\begin{equation}
mv_{\mathrm{IoU}} =
\frac{1}{|\Omega|}
\sum_{k\in\Omega}
\mathrm{IoU}(\hat{b}_k,b_k^{\ast}),
\qquad
\Omega=[\hat{t}_s,\hat{t}_e]\cap[t_s^{\ast},t_e^{\ast}].
\end{equation}
If the temporal intersection is empty, the spatial reward is set to zero. This term encourages accurate localization, stable identity tracking, and motion-aware alignment after the model has found the correct event region. Together, $t_{\mathrm{IoU}}$ and $mv_{\mathrm{IoU}}$ score the two coupled parts of STVG without collapsing temporal and spatial errors into a single signal.

\paragraph{GRPO-style optimization.}
We initialize RL from the SFT model. For each prompt $x$, we sample a group of responses $\{y_i\}_{i=1}^{G}$ and compute normalized group advantages $\hat{A}_i$ from their verifier rewards. Let
\begin{equation}
r_i(\theta)=\frac{\pi_\theta(y_i\mid x)}{\pi_{\theta_{\text{old}}}(y_i\mid x)}.
\end{equation}
We optimize the clipped objective
\begin{equation}
\mathcal{J}_{\text{RL}}(\theta)=
\mathbb{E}\left[
\min\!\left(
r_i(\theta)\hat{A}_i,
\mathrm{clip}_{[1-\epsilon_{\text{low}},\,1+\epsilon_{\text{high}}]}
\!\left(r_i(\theta)\right)\hat{A}_i
\right)
\right].
\end{equation}
We use asymmetric clipping with $\epsilon_{\text{low}}=0.20$ and $\epsilon_{\text{high}}=0.28$. The larger upper range allows beneficial updates to move slightly more when the verifier strongly prefers an output, while the lower range keeps harmful probability decreases conservative. We do not add an explicit KL penalty.

\subsection{Inference with Second-Level Tracking and Smoothing}
At inference time, the model receives the language query and second-level video units, then outputs a temporal span and a compact second-level trajectory. We parse the structured answer, discard invalid states, and apply inter-second smoothing to reduce local jitter. The smoothing step uses neighboring valid boxes to stabilize abrupt changes while preserving the predicted temporal boundaries. The final output can be kept at second-level granularity or expanded back to frame-level boxes through interpolation when required by a benchmark protocol. This inference procedure keeps the computational footprint close to the number of seconds rather than the number of frames, which is the main source of efficiency in long videos.

\section{Experiments}

\subsection{Experimental Setup}
\paragraph{Benchmarks.}
We evaluate on two standard STVG leaderboards. \textbf{VidSTG} \citep{zhang2020doesexistspatiotemporalvideo} is built from the VidOR \citep{shang2019annotating} video relation dataset and asks models to ground multi-form language queries in untrimmed videos. It contains both declarative and interrogative sentence forms, making it a useful benchmark for testing whether a model can handle different linguistic expressions of the same grounding problem. \textbf{HC-STVG} focuses on human-centric spatio-temporal grounding in complex multi-person scenes. Compared with generic object grounding, HC-STVG stresses target disambiguation, interaction understanding, and identity preservation under crowded motion. Together, the two benchmarks cover both general video relation grounding and human-centric grounding, which are the main evaluation settings for STVG. We also evaluate on \textbf{Video-MME-v2}~\citep{fu2026videommev2stagebenchmarkscomprehensive} to check whether the training pipeline preserves and improves general video understanding rather than overfitting to coordinate prediction.

\paragraph{Metrics.}
Following standard STVG evaluation, we report temporal localization and tube localization metrics. $m_{t\mathrm{IoU}}$ measures the mean temporal IoU between predicted and ground-truth time windows. $m_{v\mathrm{IoU}}$ measures the mean spatio-temporal tube IoU and is the primary indicator of full grounding quality. We also report $v\mathrm{IoU}@R$, which measures the percentage of examples whose tube IoU exceeds threshold $R$ (e.g., $0.3$ or $0.5$).
\paragraph{Compared models.}
We compare against five general MLLM baselines: \qwen{}, \seed{}, \kimi{} \citep{kimiteam2026kimik25visualagentic}, \geminiflash{}, and \geminipro{}. We also report three versions of our model: the SFT checkpoint, an RL checkpoint optimized with $mv_{\mathrm{IoU}}$ only, and the final RL checkpoint optimized with $t_{\mathrm{IoU}}+mv_{\mathrm{IoU}}$. This setup separates the effect of task-specific SFT from the additional gains brought by verifier-driven RL.

\subsection{Main Results}
Table~\ref{tab:main_results} summarizes the main STVG comparison. Our final RL model achieves the strongest grounding performance on both VidSTG and HC-STVG, reaching the best results among all evaluated systems. This is notable because our model is a 9B task-aligned checkpoint, while several baselines are substantially larger general-purpose MLLMs. The comparison suggests that STVG performance is not determined only by backbone scale: second-level formulation, grounding-and-tracking CPT, corrected reasoning supervision, and verifier-driven RL are all important for turning video-language understanding into precise trajectories.

The progression from Ours-SFT to the two RL variants further isolates the effect of alignment. SFT already adapts the model to the required output format and reasoning behavior, but RL improves the final grounding metrics by optimizing completed predictions with task-level rewards. The full verifier, $t_{\mathrm{IoU}}+mv_{\mathrm{IoU}}$, is designed to improve both temporal localization and motion-aware spatial tracking, which explains why the final RL checkpoint is expected to perform best on the full tube metrics rather than only on temporal overlap. In training, the combined $t_{\mathrm{IoU}}+mv_{\mathrm{IoU}}$ reward also converges faster than using $mv_{\mathrm{IoU}}$ alone, because the temporal component provides a denser early signal before the model has learned to predict fully accurate trajectories.
For the final RL model, we report the mean over three evaluation runs on VidSTG and HC-STVG.

% \begin{table}[t]
% \centering
% \caption{Main results on VidSTG and HC-STVG leaderboards.}
% \resizebox{\linewidth}{!}{
% \begin{tabular}{llcccc}
% \toprule
% Benchmark & Model & Params & $m_{t\mathrm{IoU}}\uparrow$ & $m_{v\mathrm{IoU}}\uparrow$ & $v\mathrm{IoU}@0.5\uparrow$ \\
% \midrule
% VidSTG & \qwen{} & 397B & 27.88 & 15.32 & TBD \\
% VidSTG & \seed{} & undisclosed & 26.28 & 16.66 &  \\
% VidSTG & \kimi{} & 1T & 43.9 & 6.42 & TBD \\
% VidSTG & \geminiflash{} & undisclosed & 8.90 & 8.90 & 4.11 \\
% VidSTG & \geminipro{} & undisclosed & 20.07 & 11.57 & 4.98 \\
% VidSTG & Ours-SFT & 9B & TBD & TBD & TBD \\
% VidSTG & Ours-RL  & 9B & 31.35 & 23.82 & 19.49 \\
% \midrule
% HC-STVG & \qwen{} & 397B & TBD & TBD & TBD \\
% HC-STVG & \seed{} & undisclosed & 42.17 & 26.68 & 9.37 \\
% HC-STVG & \kimi{} & 1T & 43.90 & 10.99 & TBD \\
% HC-STVG & \geminiflash{} & undisclosed & TBD & TBD & TBD \\
% HC-STVG & \geminipro{} & undisclosed & TBD & TBD & TBD \\
% HC-STVG & Ours-SFT & 9B & TBD & 33.75 & TBD \\
% HC-STVG & Ours-RL  & 9B & TBD & 40.99 & TBD \\

% \bottomrule
% \end{tabular}}
% \label{tab:main_results}
% \end{table}
\begin{table}[t]
\centering
\caption{Main results on VidSTG and HC-STVG leaderboards. The Ours-RL numbers are means over three evaluation runs.}
\resizebox{0.8\linewidth}{!}{%
\begin{tabular}{l|lcccc}
\toprule
Benchmark & Model & Params & $m_{t\mathrm{IoU}}\uparrow$ & $m_{v\mathrm{IoU}}\uparrow$ & $v\mathrm{IoU}@0.5\uparrow$ \\
\midrule
\multirow{7}{*}{VidSTG} & \qwen{} & 397B & 27.88 & 15.32 & 15.50 \\
                         & \seed{} & undisclosed & 26.28 & 16.66 & 9.78 \\
                         & \kimi{} & 1T & 23.99 & 16.05 & 10.46 \\
                         & \geminiflash{} & undisclosed & 18.76& 8.90 & 4.11 \\
                         & \geminipro{} & undisclosed & 20.07 & 11.57 & 4.98 \\
                         & Ours-SFT & 9B & 27.66 & 19.47 & 11.86 \\
                         & Ours-RL  & 9B & \textbf{31.35} & \textbf{23.79} & \textbf{19.49} \\
\midrule
\multirow{7}{*}{HC-STVG} & \qwen{} & 397B & 50.99 & 28.20 & 16.18 \\
                         & \seed{} & undisclosed & 42.17 & 26.68 & 9.37 \\
                         & \kimi{} & 1T & 43.9 & 10.99 & 9.04 \\
                         & \geminiflash{} & undisclosed & TBD & 16.64 & 2.84 \\
                         & \geminipro{} & undisclosed & 43.81 & 20.24 & 4.82 \\
                         & Ours-SFT & 9B & 51.23 & 33.75 & 26.18 \\
                         & Ours-RL  & 9B & \textbf{60.04} & \textbf{40.99} & \textbf{34.12} \\
\bottomrule
\end{tabular}%
}
\label{tab:main_results}
\end{table}

\begin{figure}[t]
\centering
\resizebox{\linewidth}{!}{%
\begin{tikzpicture}[x=1cm,y=1cm]
    \definecolor{baselinebar}{RGB}{116,139,171}
    \definecolor{oursbar}{RGB}{36,132,97}
    \definecolor{panelbg}{RGB}{248,249,251}
    \definecolor{panelrule}{RGB}{210,216,224}

    \fill[panelbg] (0,0) rectangle (12.8,6.1);
    \draw[panelrule] (0,0) rectangle (12.8,6.1);
    \draw[panelrule] (0,3.15) -- (12.8,3.15);
    \draw[panelrule] (6.25,0) -- (6.25,3.15);

    \node[font=\small\bfseries,anchor=west] at (0.35,5.72) {Video-MME-v2};
    \fill[baselinebar] (3.05,5.72) circle (0.06);
    \node[font=\scriptsize,anchor=west] at (3.18,5.72) {GLM-4.1V-Thinking};
    \fill[oursbar] (6.15,5.72) circle (0.06);
    \node[font=\scriptsize,anchor=west] at (6.28,5.72) {Ours};
    \draw[black!45] (2.25,3.75) -- (11.75,3.75);
    \foreach \tick in {0,10,20,30} {
        \draw[black!18] ({2.25+\tick*0.3167},3.75) -- ({2.25+\tick*0.3167},5.35);
        \node[font=\scriptsize,anchor=north] at ({2.25+\tick*0.3167},3.62) {\tick};
    }
    \node[font=\scriptsize,anchor=east] at (2.0,4.55) {Non-linear};
    \node[font=\scriptsize,anchor=east] at (2.0,5.15) {Simple Acc.};
    \draw[black!35,line width=0.8pt] ({2.25+9.4*0.3167},4.55) -- ({2.25+12.6*0.3167},4.55);
    \draw[black!35,line width=0.8pt] ({2.25+23.7*0.3167},5.15) -- ({2.25+28.0*0.3167},5.15);
    \fill[baselinebar] ({2.25+9.4*0.3167},4.55) circle (0.07);
    \fill[oursbar] ({2.25+12.6*0.3167},4.55) circle (0.07);
    \fill[baselinebar] ({2.25+23.7*0.3167},5.15) circle (0.07);
    \fill[oursbar] ({2.25+28.0*0.3167},5.15) circle (0.07);
    \node[font=\scriptsize,anchor=south] at ({2.25+9.4*0.3167},4.64) {9.4};
    \node[font=\scriptsize\bfseries,anchor=south] at ({2.25+12.6*0.3167},4.64) {12.6};
    \node[font=\scriptsize,anchor=south] at ({2.25+23.7*0.3167},5.24) {23.7};
    \node[font=\scriptsize\bfseries,anchor=south] at ({2.25+28.0*0.3167},5.24) {28.0};

    \node[font=\small\bfseries,anchor=west] at (0.35,2.78) {FPS Sensitivity};
    \draw[black!40] (0.65,0.75) -- (5.75,0.75);
    \foreach \x/\lab/\val/\height/\bold in {1.25/$0.5$ FPS/31.4/1.43/0,3.15/$1$ FPS/39.08/1.78/1,5.05/Dynamic/38.5/1.75/0} {
        \fill[baselinebar!55] (\x-0.23,0.75) rectangle (\x+0.23,{0.75+\height});
        \node[font=\scriptsize,anchor=north] at (\x,0.55) {\lab};
        \ifnum\bold=1
            \node[font=\scriptsize\bfseries,anchor=south] at (\x,{0.85+\height}) {\val};
        \else
            \node[font=\scriptsize,anchor=south] at (\x,{0.85+\height}) {\val};
        \fi
    }

    \node[font=\small\bfseries,anchor=west] at (6.65,2.78) {RL Reward Ablation};
    \draw[black!45] (6.65,2.25) -- (12.35,2.25);
    \draw[black!25] (6.65,1.82) -- (12.35,1.82);
    \draw[black!45] (6.65,0.50) -- (12.35,0.50);
    \node[font=\scriptsize,anchor=west] at (6.8,2.03) {Reward};
    \node[font=\scriptsize] at (10.15,2.03) {$m_{t\mathrm{IoU}}\uparrow$};
    \node[font=\scriptsize] at (11.75,2.03) {$m_{v\mathrm{IoU}}\uparrow$};
    \node[font=\scriptsize,anchor=west] at (6.8,1.40) {$mv_{\mathrm{IoU}}$ only};
    \node[font=\scriptsize,anchor=west] at (6.8,0.84) {$t_{\mathrm{IoU}}+mv_{\mathrm{IoU}}$};
    \node[font=\scriptsize] at (10.15,1.40) {29.11};
    \node[font=\scriptsize] at (11.75,1.40) {22.59};
    \node[font=\scriptsize\bfseries] at (10.15,0.84) {31.35};
    \node[font=\scriptsize\bfseries] at (11.75,0.84) {23.82};
\end{tikzpicture}%
}
\vspace{-0.4em}
\caption{Compact auxiliary results and ablations. Left: Video-MME-v2 head-to-head comparison. Right: training FPS sensitivity, tested on Dancetrack \citep{sun2022dancetrackmultiobjecttrackinguniform} and the RL reward ablation.}
\label{fig:aux_dashboard}
\end{figure}
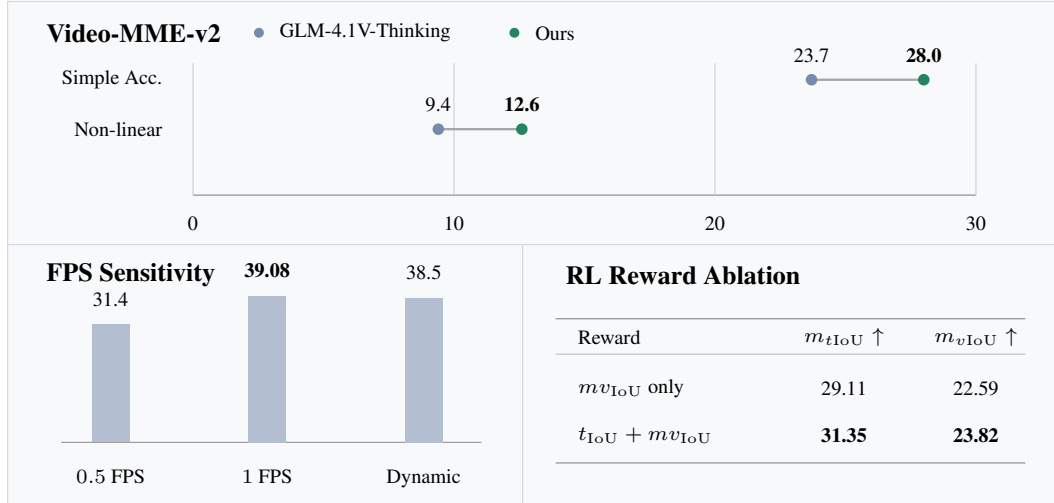

Beyond STVG, we also compare on Video-MME-v2 to assess general video understanding. This comparison is important because our training pipeline includes tracking-heavy data and coordinate supervision, which could in principle narrow the model toward localization-only behavior. Figure~\ref{fig:aux_dashboard} shows the opposite trend: our model improves over GLM-4.1V-Thinking on both non-linear reasoning accuracy and simple accuracy, reaching $12.6$ and $28.0$ compared with the baseline's $9.4$ and $23.7$. This indicates that the spatio-temporal grounding task can successfully feed back into general video understanding, rather than merely producing a task-specific localization model. It also supports our design choice of including general video understanding data in CPT, so that the model can combine precise grounding with broader event and commonsense reasoning.

\subsection{Training FPS Sensitivity}
We evaluate how the training frame rate affects grounding quality. Lower FPS reduces training cost but may miss short transitions, while higher FPS exposes denser temporal evidence at the cost of longer visual sequences. We also include a dynamic-FPS setting that adjusts the sampling density according to the video or event duration. Figure~\ref{fig:aux_dashboard} reports the sensitivity study; 1 FPS achieves the best result among the evaluated training frame-rate settings.

\section{Discussion}
\subsection{Ablation Studies}
We study the reward design in Figure~\ref{fig:aux_dashboard} by comparing an RL variant optimized only with the motion-aware tube reward $mv_{\mathrm{IoU}}$ against our verifier $t_{\mathrm{IoU}}+mv_{\mathrm{IoU}}$, while keeping the model, data, second-level formulation, and training pipeline fixed. The $mv_{\mathrm{IoU}}$-only variant directly optimizes the final tube quality, but it can only provide a weak learning signal early in RL. When the predicted temporal window is far from the ground truth, the reward becomes close to zero even if the model has partially identified the correct target. Adding $t_{\mathrm{IoU}}$ makes the reward denser and leads to faster convergence: the policy can first learn to move the event window toward the correct segment, after which $mv_{\mathrm{IoU}}$ becomes meaningful for refining the trajectory.

\subsection{Case Analysis and Generalization}
\begin{figure}[t]
    \centering
    \begin{minipage}[t]{0.5\linewidth}
        \centering
        \parbox[t][0.255\textheight][t]{\linewidth}{\centering
        \includegraphics[width=\linewidth,trim=18 14 18 14,clip]{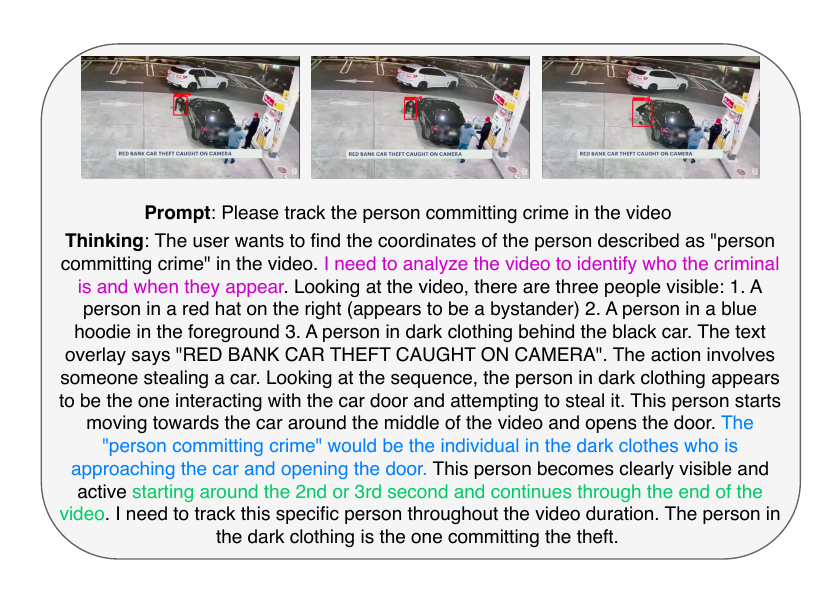}}
    \end{minipage}%
    \begin{minipage}[t]{0.5\linewidth}
        \centering
        \parbox[t][0.255\textheight][t]{\linewidth}{\centering
        \includegraphics[width=\linewidth,trim=18 14 18 14,clip]{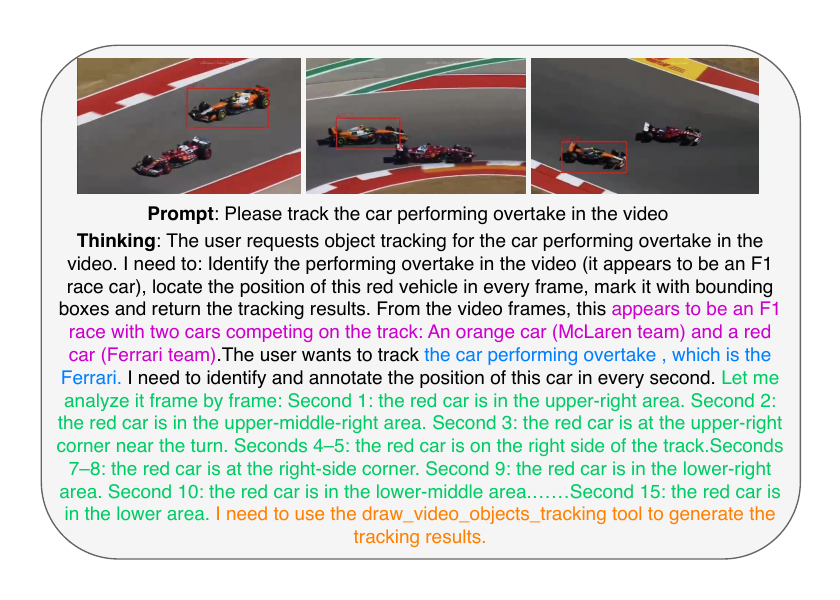}}
    \end{minipage}
    \caption{Qualitative case studies. Left: the model localizes the queried theft-related event and tracks the referred target despite nearby distractors. Right: the model keeps the target identity stable across motion and visually similar nearby objects.}
    \label{fig:qualitative_cases}
\end{figure}

Qualitatively, the strongest gains appear in long videos with distractors and queries whose target cannot be identified from a category label alone. Frame-level or generic MLLM baselines often locate the correct object in a few frames but fail to maintain a stable tube, or they choose a temporal window that includes large amounts of irrelevant context. Our model more consistently identifies the queried target before producing the final trajectory, which reduces identity switches. Figure~\ref{fig:qualitative_cases} provides two representative case studies.

In the left panel of Figure~\ref{fig:qualitative_cases}, the query asks the model to track ``the person committing crime,'' which requires understanding the event rather than simply matching an object category. The video contains multiple people: a person in a red hat near the right side, a person in a blue hoodie in the foreground, and a person in dark clothing around the black car. The model's reasoning first decomposes the scene, compares these candidate people, uses the video context and the on-screen cue about car theft, and then identifies the person in dark clothing as the one approaching the car and opening the door. This reasoning trace gives a useful high-level guide before coordinate prediction: it resolves who should be tracked and when the relevant action begins, after which the model produces a precise trajectory for the correct person. This case shows that the method preserves strong general video understanding while making it actionable for spatio-temporal tracking.

In the right panel of Figure~\ref{fig:qualitative_cases}, the prompt asks for the car performing the overtake. This is difficult for a traditional localization or tracking model without large-scale CPT, because the model must understand a motorsport scene, infer that the orange car belongs to McLaren and the red car belongs to Ferrari, and decide which vehicle is performing the overtaking maneuver. Our method uses the MLLM's world knowledge to connect visual appearance, team identity, and race dynamics before emitting the tracking boxes. This example illustrates an important generalization advantage: when the query depends on domain knowledge rather than only visual similarity, the MLLM-based formulation can identify targets that conventional pipelines are unlikely to distinguish reliably.

\section{Conclusion}
We present a practical MLLM framework for STVG. The method uses second-level tracking to reduce long context, corrected reasoning SFT to teach target and temporal reasoning, and verifier-driven RL to directly optimize numeric grounding quality. Experiments on VidSTG and HC-STVG show that a 9B task-aligned model can outperform much larger general MLLM baselines, suggesting that representation, supervision, and reward design are central for video grounding. The Video-MME-v2 result further shows that localization-oriented training can feed back into general video understanding. 

\paragraph{Limitations.}
Future work can use the same MLLM reasoning ability to assist broader video understanding tasks, such as event parsing, long-horizon activity analysis, and interactive video agents. Besides, extending the method to noisier open-world videos or real-time settings will require stronger filtering, uncertainty handling, and adaptive temporal resolution.

% Limitations section intentionally left empty in this arXiv source package.

\bibliographystyle{plainnat}
\bibliography{references}

@misc{fu2026videommev2stagebenchmarkscomprehensive,
  title={Video-MME-v2: Towards the Next Stage in Benchmarks for Comprehensive Video Understanding},
  author={Chaoyou Fu and Haozhi Yuan and Yuhao Dong and Yi-Fan Zhang and Yunhang Shen and Xiaoxing Hu and Xueying Li and Jinsen Su and Chengwu Long and Xiaoyao Xie and Yongkang Xie and Xiawu Zheng and Xue Yang and Haoyu Cao and Yunsheng Wu and Ziwei Liu and Xing Sun and Caifeng Shan and Ran He},
  year={2026},
  eprint={2604.05015},
  archivePrefix={arXiv},
  primaryClass={cs.CV},
  url={https://arxiv.org/abs/2604.05015}
}

@misc{chen2024ictimageobjectcrossleveltrusted,
      title={ICT: Image-Object Cross-Level Trusted Intervention for Mitigating Object Hallucination in Large Vision-Language Models}, 
      author={Junzhe Chen and Tianshu Zhang and Shiyu Huang and Yuwei Niu and Linfeng Zhang and Lijie Wen and Xuming Hu},
      year={2024},
      eprint={2411.15268},
      archivePrefix={arXiv},
      primaryClass={cs.CV},
      url={https://arxiv.org/abs/2411.15268}, 
}

@inproceedings{shang2019annotating,
    title={Annotating Objects and Relations in User-Generated Videos},
    author={Shang, Xindi and Di, Donglin and Xiao, Junbin and Cao, Yu and Yang, Xun and Chua, Tat-Seng},
    booktitle={Proceedings of the 2019 on International Conference on Multimedia Retrieval},
    pages={279--287},
    year={2019},
    organization={ACM}
}

@misc{ravi2024sam2segmentimages,
      title={SAM 2: Segment Anything in Images and Videos}, 
      author={Nikhila Ravi and Valentin Gabeur and Yuan-Ting Hu and Ronghang Hu and Chaitanya Ryali and Tengyu Ma and Haitham Khedr and Roman Rädle and Chloe Rolland and Laura Gustafson and Eric Mintun and Junting Pan and Kalyan Vasudev Alwala and Nicolas Carion and Chao-Yuan Wu and Ross Girshick and Piotr Dollár and Christoph Feichtenhofer},
      year={2024},
      eprint={2408.00714},
      archivePrefix={arXiv},
      primaryClass={cs.CV},
      url={https://arxiv.org/abs/2408.00714}, 
}

@misc{fan2020lasothighqualitylargescalesingle,
      title={LaSOT: A High-quality Large-scale Single Object Tracking Benchmark}, 
      author={Heng Fan and Hexin Bai and Liting Lin and Fan Yang and Peng Chu and Ge Deng and Sijia Yu and Harshit and Mingzhen Huang and Juehuan Liu and Yong Xu and Chunyuan Liao and Lin Yuan and Haibin Ling},
      year={2020},
      eprint={2009.03465},
      archivePrefix={arXiv},
      primaryClass={cs.CV},
      url={https://arxiv.org/abs/2009.03465}, 
}

@article{Huang_2021,
   title={GOT-10k: A Large High-Diversity Benchmark for Generic Object Tracking in the Wild},
   volume={43},
   ISSN={1939-3539},
   url={http://dx.doi.org/10.1109/TPAMI.2019.2957464},
   DOI={10.1109/tpami.2019.2957464},
   number={5},
   journal={IEEE Transactions on Pattern Analysis and Machine Intelligence},
   publisher={Institute of Electrical and Electronics Engineers (IEEE)},
   author={Huang, Lianghua and Zhao, Xin and Huang, Kaiqi},
   year={2021},
   month=May, pages={1562–1577} }

@inproceedings{miao2021vspw,
  title={Vspw: A large-scale dataset for video scene parsing in the wild},
  author={Miao, Jiaxu and Wei, Yunchao and Wu, Yu and Liang, Chen and Li, Guangrui and Yang, Yi},
  booktitle={Proceedings of the IEEE/CVF Conference on Computer Vision and Pattern Recognition},
  pages={4133--4143},
  year={2021}
}

@misc{cui2023sportsmotlargemultiobjecttracking,
      title={SportsMOT: A Large Multi-Object Tracking Dataset in Multiple Sports Scenes}, 
      author={Yutao Cui and Chenkai Zeng and Xiaoyu Zhao and Yichun Yang and Gangshan Wu and Limin Wang},
      year={2023},
      eprint={2304.05170},
      archivePrefix={arXiv},
      primaryClass={cs.CV},
      url={https://arxiv.org/abs/2304.05170}, 
}

@misc{müller2018trackingnetlargescaledatasetbenchmark,
      title={TrackingNet: A Large-Scale Dataset and Benchmark for Object Tracking in the Wild}, 
      author={Matthias Müller and Adel Bibi and Silvio Giancola and Salman Al-Subaihi and Bernard Ghanem},
      year={2018},
      eprint={1803.10794},
      archivePrefix={arXiv},
      primaryClass={cs.CV},
      url={https://arxiv.org/abs/1803.10794}, 
}

@misc{zhang2020doesexistspatiotemporalvideo,
      title={Where Does It Exist: Spatio-Temporal Video Grounding for Multi-Form Sentences}, 
      author={Zhu Zhang and Zhou Zhao and Yang Zhao and Qi Wang and Huasheng Liu and Lianli Gao},
      year={2020},
      eprint={2001.06891},
      archivePrefix={arXiv},
      primaryClass={cs.CV},
      url={https://arxiv.org/abs/2001.06891}, 
}

@misc{sun2022dancetrackmultiobjecttrackinguniform,
      title={DanceTrack: Multi-Object Tracking in Uniform Appearance and Diverse Motion}, 
      author={Peize Sun and Jinkun Cao and Yi Jiang and Zehuan Yuan and Song Bai and Kris Kitani and Ping Luo},
      year={2022},
      eprint={2111.14690},
      archivePrefix={arXiv},
      primaryClass={cs.CV},
      url={https://arxiv.org/abs/2111.14690}, 
}

@misc{li2025llavastmultimodallargelanguage,
      title={LLaVA-ST: A Multimodal Large Language Model for Fine-Grained Spatial-Temporal Understanding}, 
      author={Hongyu Li and Jinyu Chen and Ziyu Wei and Shaofei Huang and Tianrui Hui and Jialin Gao and Xiaoming Wei and Si Liu},
      year={2025},
      eprint={2501.08282},
      archivePrefix={arXiv},
      primaryClass={cs.CV},
      url={https://arxiv.org/abs/2501.08282}, 
}

@misc{wang2025spacevllmendowingmultimodallarge,
      title={SpaceVLLM: Endowing Multimodal Large Language Model with Spatio-Temporal Video Grounding Capability}, 
      author={Jiankang Wang and Zhihan Zhang and Zhihang Liu and Yang Li and Jiannan Ge and Hongtao Xie and Yongdong Zhang},
      year={2025},
      eprint={2503.13983},
      archivePrefix={arXiv},
      primaryClass={cs.CV},
      url={https://arxiv.org/abs/2503.13983}, 
}

@misc{bai2023qwenvlversatilevisionlanguagemodel,
      title={Qwen-VL: A Versatile Vision-Language Model for Understanding, Localization, Text Reading, and Beyond}, 
      author={Jinze Bai and Shuai Bai and Shusheng Yang and Shijie Wang and Sinan Tan and Peng Wang and Junyang Lin and Chang Zhou and Jingren Zhou},
      year={2023},
      eprint={2308.12966},
      archivePrefix={arXiv},
      primaryClass={cs.CV},
      url={https://arxiv.org/abs/2308.12966}, 
}

@misc{ren2024timechattimesensitivemultimodallarge,
      title={TimeChat: A Time-sensitive Multimodal Large Language Model for Long Video Understanding}, 
      author={Shuhuai Ren and Linli Yao and Shicheng Li and Xu Sun and Lu Hou},
      year={2024},
      eprint={2312.02051},
      archivePrefix={arXiv},
      primaryClass={cs.CV},
      url={https://arxiv.org/abs/2312.02051}, 
}

@misc{pramanick2025enrichdetectvideotemporal,
      title={Enrich and Detect: Video Temporal Grounding with Multimodal LLMs}, 
      author={Shraman Pramanick and Effrosyni Mavroudi and Yale Song and Rama Chellappa and Lorenzo Torresani and Triantafyllos Afouras},
      year={2025},
      eprint={2510.17023},
      archivePrefix={arXiv},
      primaryClass={cs.CV},
      url={https://arxiv.org/abs/2510.17023}, 
}

@misc{vteam2026glm45vglm41vthinkingversatilemultimodal,
      title={GLM-4.5V and GLM-4.1V-Thinking: Towards Versatile Multimodal Reasoning with Scalable Reinforcement Learning}, 
      author={V Team and Wenyi Hong and Wenmeng Yu and Xiaotao Gu and Guo Wang and Guobing Gan and Haomiao Tang and Jiale Cheng and Ji Qi and Junhui Ji and Lihang Pan and Shuaiqi Duan and Weihan Wang and Yan Wang and Yean Cheng and Zehai He and Zhe Su and Zhen Yang and Ziyang Pan and Aohan Zeng and Baoxu Wang and Bin Chen and Boyan Shi and Changyu Pang and Chenhui Zhang and Da Yin and Fan Yang and Guoqing Chen and Haochen Li and Jiale Zhu and Jiali Chen and Jiaxing Xu and Jiazheng Xu and Jing Chen and Jinghao Lin and Jinhao Chen and Jinjiang Wang and Junjie Chen and Leqi Lei and Letian Gong and Leyi Pan and Mingdao Liu and Mingde Xu and Mingzhi Zhang and Qinkai Zheng and Ruiliang Lyu and Shangqin Tu and Sheng Yang and Shengbiao Meng and Shi Zhong and Shiyu Huang and Shuyuan Zhao and Siyan Xue and Tianshu Zhang and Tianwei Luo and Tianxiang Hao and Tianyu Tong and Wei Jia and Wenkai Li and Xiao Liu and Xiaohan Zhang and Xin Lyu and Xinyu Zhang and Xinyue Fan and Xuancheng Huang and Yadong Xue and Yanfeng Wang and Yanling Wang and Yanzi Wang and Yifan An and Yifan Du and Yiheng Huang and Yilin Niu and Yiming Shi and Yu Wang and Yuan Wang and Yuanchang Yue and Yuchen Li and Yusen Liu and Yutao Zhang and Yuting Wang and Yuxuan Zhang and Zhao Xue and Zhengxiao Du and Zhenyu Hou and Zihan Wang and Peng Zhang and Debing Liu and Bin Xu and Juanzi Li and Minlie Huang and Yuxiao Dong and Jie Tang},
      year={2026},
      eprint={2507.01006},
      archivePrefix={arXiv},
      primaryClass={cs.CV},
      url={https://arxiv.org/abs/2507.01006}, 
}

@misc{bai2025qwen3vltechnicalreport,
      title={Qwen3-VL Technical Report}, 
      author={Shuai Bai and Yuxuan Cai and Ruizhe Chen and Keqin Chen and Xionghui Chen and Zesen Cheng and Lianghao Deng and Wei Ding and Chang Gao and Chunjiang Ge and Wenbin Ge and Zhifang Guo and Qidong Huang and Jie Huang and Fei Huang and Binyuan Hui and Shutong Jiang and Zhaohai Li and Mingsheng Li and Mei Li and Kaixin Li and Zicheng Lin and Junyang Lin and Xuejing Liu and Jiawei Liu and Chenglong Liu and Yang Liu and Dayiheng Liu and Shixuan Liu and Dunjie Lu and Ruilin Luo and Chenxu Lv and Rui Men and Lingchen Meng and Xuancheng Ren and Xingzhang Ren and Sibo Song and Yuchong Sun and Jun Tang and Jianhong Tu and Jianqiang Wan and Peng Wang and Pengfei Wang and Qiuyue Wang and Yuxuan Wang and Tianbao Xie and Yiheng Xu and Haiyang Xu and Jin Xu and Zhibo Yang and Mingkun Yang and Jianxin Yang and An Yang and Bowen Yu and Fei Zhang and Hang Zhang and Xi Zhang and Bo Zheng and Humen Zhong and Jingren Zhou and Fan Zhou and Jing Zhou and Yuanzhi Zhu and Ke Zhu},
      year={2025},
      eprint={2511.21631},
      archivePrefix={arXiv},
      primaryClass={cs.CV},
      url={https://arxiv.org/abs/2511.21631}, 
}

@misc{vteam2026glm5vturbonativefoundationmodel,
      title={GLM-5V-Turbo: Toward a Native Foundation Model for Multimodal Agents}, 
      author={V Team and Wenyi Hong and Xiaotao Gu and Ziyang Pan and Zhen Yang and Yuting Wang and Yue Wang and Yuanchang Yue and Yu Wang and Yanling Wang and Yan Wang and Xijun Liu and Wenmeng Yu and Weihan Wang and Wei Li and Shuaiqi Duan and Sheng Yang and Ruiliang Lv and Mingdao Liu and Lihang Pan and Ke Ning and Junhui Ji and Jinjiang Wang and Jing Chen and Jiazheng Xu and Jiale Zhu and Jiale Cheng and Ji Qi and Guobing Gan and Guo Wang and Cong Yao and Zijun Dou and Zihao Zhou and Zihan Wang and Zhiqi Ge and Zhijie Li and Zhenyu Hou and Zhao Xue and Zehui Wang and Zehan Qi and Zehai He and Yutao Zhang and Yusen Liu and Yukuo Cen and Yuchen Li and Yuan Wang and Yu Yang and Yongbin Liu and Yijian Lu and Yifan Xu and Yanzi Wang and Yanxiao Zhao and Yanfeng Wang and Yadong Xue and Yabo Xu and Xinyu Zhang and Xinyu Liu and Xiao Liu and Wenyi Zhao and Wenkai Li and Tianyu Tong and Tianshu Zhang and Shudan Zhang and Shengdong Yan and Qinkai Zheng and Mingde Xu and Licheng Bao and lat Long long and Jiaxing Xu and Jiaxin Fan and Jiawen Qian and Jiali Chen and Jiahui Lin and Jiadai Sun and Haozhi Zheng and Haoran Wang and Haochen Li and Hanyu Liu and Han Xu and Fan Yang and Dan Zhang and Da Yin and Chuangxin Zhao and Chengcheng Wu and Boyan Shi and Bowen Lv and Bowei Jia and Bo Li and Bin Chen and Baoxu Wang and Peng Zhang and Debing Liu and Bin Xu and Juanzi Li and Minlie Huang and Yuxiao Dong and Jie Tang},
      year={2026},
      eprint={2604.26752},
      archivePrefix={arXiv},
      primaryClass={cs.CV},
      url={https://arxiv.org/abs/2604.26752}, 
}

@misc{gu2025thinkingboundingboxesenhancing,
      title={Thinking With Bounding Boxes: Enhancing Spatio-Temporal Video Grounding via Reinforcement Fine-Tuning}, 
      author={Xin Gu and Haoji Zhang and Qihang Fan and Jingxuan Niu and Zhipeng Zhang and Libo Zhang and Guang Chen and Fan Chen and Longyin Wen and Sijie Zhu},
      year={2025},
      eprint={2511.21375},
      archivePrefix={arXiv},
      primaryClass={cs.CV},
      url={https://arxiv.org/abs/2511.21375}, 
}

@misc{zhang2020objectawaremultibranchrelationnetworks,
      title={Object-Aware Multi-Branch Relation Networks for Spatio-Temporal Video Grounding}, 
      author={Zhu Zhang and Zhou Zhao and Zhijie Lin and Baoxing Huai and Nicholas Jing Yuan},
      year={2020},
      eprint={2008.06941},
      archivePrefix={arXiv},
      primaryClass={cs.CV},
      url={https://arxiv.org/abs/2008.06941}, 
}

@misc{gu2025knowingtargettargetawaretransformer,
      title={Knowing Your Target: Target-Aware Transformer Makes Better Spatio-Temporal Video Grounding}, 
      author={Xin Gu and Yaojie Shen and Chenxi Luo and Tiejian Luo and Yan Huang and Yuewei Lin and Heng Fan and Libo Zhang},
      year={2025},
      eprint={2502.11168},
      archivePrefix={arXiv},
      primaryClass={cs.CV},
      url={https://arxiv.org/abs/2502.11168}, 
}

@InProceedings{Su_2021_ICCV,
    author    = {Su, Rui and Yu, Qian and Xu, Dong},
    title     = {STVGBert: A Visual-Linguistic Transformer Based Framework for Spatio-Temporal Video Grounding},
    booktitle = {Proceedings of the IEEE/CVF International Conference on Computer Vision (ICCV)},
    month     = {October},
    year      = {2021},
    pages     = {1533-1542}
}

@misc{yang2025unleashingpotentialmultimodalllms,
      title={Unleashing the Potential of Multimodal LLMs for Zero-Shot Spatio-Temporal Video Grounding}, 
      author={Zaiquan Yang and Yuhao Liu and Gerhard Hancke and Rynson W. H. Lau},
      year={2025},
      eprint={2509.15178},
      archivePrefix={arXiv},
      primaryClass={cs.CV},
      url={https://arxiv.org/abs/2509.15178}, 
}

@misc{fu2025omniptunleashingpotentiallarge,
      title={OmniPT: Unleashing the Potential of Large Vision Language Models for Pedestrian Tracking and Understanding}, 
      author={Teng Fu and Mengyang Zhao and Ke Niu and Kaixin Peng and Bin Li},
      year={2025},
      eprint={2511.17053},
      archivePrefix={arXiv},
      primaryClass={cs.CV},
      url={https://arxiv.org/abs/2511.17053}, 
}

@misc{yang2026tracevisiontrajectoryawarevisionlanguagemodel,
      title={TraceVision: Trajectory-Aware Vision-Language Model for Human-Like Spatial Understanding}, 
      author={Fan Yang and Shurong Zheng and Hongyin Zhao and Yufei Zhan and Xin Li and Yousong Zhu and Chaoyang Zhao Ming Tang and Jinqiao Wang},
      year={2026},
      eprint={2602.19768},
      archivePrefix={arXiv},
      primaryClass={cs.CV},
      url={https://arxiv.org/abs/2602.19768}, 
}

@misc{gao202511,
      title={1 + 1 > 2: Detector-Empowered Video Large Language Model for Spatio-Temporal Grounding and Reasoning}, 
      author={Shida Gao and Feng Xue and Xiangfeng Wang and Anlong Ming and Teng Long and Yihua Shao and Haozhe Wang and Zhaowen Lin and Wei Wang and Nicu Sebe},
      year={2025},
      eprint={2512.06673},
      archivePrefix={arXiv},
      primaryClass={cs.CV},
      url={https://arxiv.org/abs/2512.06673}, 
}

@misc{wu2026massmotionawarespatialtemporalgrounding,
      title={MASS: Motion-Aware Spatial-Temporal Grounding for Physics Reasoning and Comprehension in Vision-Language Models}, 
      author={Xiyang Wu and Zongxia Li and Jihui Jin and Guangyao Shi and Gouthaman KV and Vishnu Raj and Nilotpal Sinha and Jingxi Chen and Fan Du and Dinesh Manocha},
      year={2026},
      eprint={2511.18373},
      archivePrefix={arXiv},
      primaryClass={cs.CV},
      url={https://arxiv.org/abs/2511.18373}, 
}

@misc{chamiti2025refergptzeroshotreferringmultiobject,
      title={ReferGPT: Towards Zero-Shot Referring Multi-Object Tracking}, 
      author={Tzoulio Chamiti and Leandro Di Bella and Adrian Munteanu and Nikos Deligiannis},
      year={2025},
      eprint={2504.09195},
      archivePrefix={arXiv},
      primaryClass={cs.CV},
      url={https://arxiv.org/abs/2504.09195}, 
}

@misc{wang2024elysiumexploringobjectlevelperception,
      title={Elysium: Exploring Object-level Perception in Videos via MLLM}, 
      author={Han Wang and Yanjie Wang and Yongjie Ye and Yuxiang Nie and Can Huang},
      year={2024},
      eprint={2403.16558},
      archivePrefix={arXiv},
      primaryClass={cs.CV},
      url={https://arxiv.org/abs/2403.16558}, 
}

@misc{munasinghe2025videoglammlargemultimodalmodel,
      title={VideoGLaMM: A Large Multimodal Model for Pixel-Level Visual Grounding in Videos}, 
      author={Shehan Munasinghe and Hanan Gani and Wenqi Zhu and Jiale Cao and Eric Xing and Fahad Shahbaz Khan and Salman Khan},
      year={2025},
      eprint={2411.04923},
      archivePrefix={arXiv},
      primaryClass={cs.CV},
      url={https://arxiv.org/abs/2411.04923}, 
}

@misc{meng2026openo3videogroundedvideoreasoning,
      title={Open-o3-Video: Grounded Video Reasoning with Explicit Spatio-Temporal Evidence}, 
      author={Jiahao Meng and Xiangtai Li and Haochen Wang and Yue Tan and Tao Zhang and Lingdong Kong and Yunhai Tong and Anran Wang and Zhiyang Teng and Yujing Wang and Zhuochen Wang},
      year={2026},
      eprint={2510.20579},
      archivePrefix={arXiv},
      primaryClass={cs.CV},
      url={https://arxiv.org/abs/2510.20579}, 
}

@misc{li2025videochatr1enhancingspatiotemporalperception,
      title={VideoChat-R1: Enhancing Spatio-Temporal Perception via Reinforcement Fine-Tuning}, 
      author={Xinhao Li and Ziang Yan and Desen Meng and Lu Dong and Xiangyu Zeng and Yinan He and Yali Wang and Yu Qiao and Yi Wang and Limin Wang},
      year={2025},
      eprint={2504.06958},
      archivePrefix={arXiv},
      primaryClass={cs.CV},
      url={https://arxiv.org/abs/2504.06958}, 
}

@misc{cheng2025vstarbenchmarkingvideollmsvideo,
      title={V-STaR: Benchmarking Video-LLMs on Video Spatio-Temporal Reasoning}, 
      author={Zixu Cheng and Jian Hu and Ziquan Liu and Chenyang Si and Wei Li and Shaogang Gong},
      year={2025},
      eprint={2503.11495},
      archivePrefix={arXiv},
      primaryClass={cs.CV},
      url={https://arxiv.org/abs/2503.11495}, 
}

@misc{meng2026videozerobenchprobinglimitsvideo,
      title={VideoZeroBench: Probing the Limits of Video MLLMs with Spatio-Temporal Evidence Verification}, 
      author={Jiahao Meng and Tan Yue and Qi Xu and Haochen Wang and Zhongwei Ren and Weisong Liu and Yuhao Wang and Renrui Zhang and Yunhai Tong and Haodong Duan},
      year={2026},
      eprint={2604.01569},
      archivePrefix={arXiv},
      primaryClass={cs.CV},
      url={https://arxiv.org/abs/2604.01569}, 
}

@misc{viditeam2026vidi25largemultimodalmodels,
      title={Vidi2.5: Large Multimodal Models for Video Understanding and Creation}, 
      author={Vidi Team and Chia-Wen Kuo and Chuang Huang and Dawei Du and Fan Chen and Fanding Lei and Feng Gao and Guang Chen and Haoji Zhang and Haojun Zhao and Jin Liu and Jingjing Zhuge and Lili Fang and Lingxi Zhang and Longyin Wen and Lu Guo and Lu Xu and Lusha Li and Qihang Fan and Rachel Deng and Shaobo Fang and Shu Zhang and Sijie Zhu and Stuart Siew and Weiyan Tao and Wen Zhong and Xiaohui Shen and Xin Gu and Ye Yuan and Yicheng He and Yiming Cui and Zhenfang Chen and Zhihua Wu and Zuhua Lin},
      year={2026},
      eprint={2511.19529},
      archivePrefix={arXiv},
      primaryClass={cs.CV},
      url={https://arxiv.org/abs/2511.19529}, 
}

@misc{clark2026molmo2openweightsdata,
      title={Molmo2: Open Weights and Data for Vision-Language Models with Video Understanding and Grounding}, 
      author={Christopher Clark and Jieyu Zhang and Zixian Ma and Jae Sung Park and Mohammadreza Salehi and Rohun Tripathi and Sangho Lee and Zhongzheng Ren and Chris Dongjoo Kim and Yinuo Yang and Vincent Shao and Yue Yang and Weikai Huang and Ziqi Gao and Taira Anderson and Jianrui Zhang and Jitesh Jain and George Stoica and Winson Han and Ali Farhadi and Ranjay Krishna},
      year={2026},
      eprint={2601.10611},
      archivePrefix={arXiv},
      primaryClass={cs.CV},
      url={https://arxiv.org/abs/2601.10611}, 
}

@misc{yang2022tubedetrspatiotemporalvideogrounding,
      title={TubeDETR: Spatio-Temporal Video Grounding with Transformers}, 
      author={Antoine Yang and Antoine Miech and Josef Sivic and Ivan Laptev and Cordelia Schmid},
      year={2022},
      eprint={2203.16434},
      archivePrefix={arXiv},
      primaryClass={cs.CV},
      url={https://arxiv.org/abs/2203.16434}, 
}

@misc{kimiteam2026kimik25visualagentic,
      title={Kimi K2.5: Visual Agentic Intelligence}, 
      author={Kimi Team and Tongtong Bai and Yifan Bai and Yiping Bao and S. H. Cai and Yuan Cao and Y. Charles and H. S. Che and Cheng Chen and Guanduo Chen and Huarong Chen and Jia Chen and Jiahao Chen and Jianlong Chen and Jun Chen and Kefan Chen and Liang Chen and Ruijue Chen and Xinhao Chen and Yanru Chen and Yanxu Chen and Yicun Chen and Yimin Chen and Yingjiang Chen and Yuankun Chen and Yujie Chen and Yutian Chen and Zhirong Chen and Ziwei Chen and Dazhi Cheng and Minghan Chu and Jialei Cui and Jiaqi Deng and Muxi Diao and Hao Ding and Mengfan Dong and Mengnan Dong and Yuxin Dong and Yuhao Dong and Angang Du and Chenzhuang Du and Dikang Du and Lingxiao Du and Yulun Du and Yu Fan and Shengjun Fang and Qiulin Feng and Yichen Feng and Garimugai Fu and Kelin Fu and Hongcheng Gao and Tong Gao and Yuyao Ge and Shangyi Geng and Chengyang Gong and Xiaochen Gong and Zhuoma Gongque and Qizheng Gu and Xinran Gu and Yicheng Gu and Longyu Guan and Yuanying Guo and Xiaoru Hao and Weiran He and Wenyang He and Yunjia He and Chao Hong and Hao Hu and Jiaxi Hu and Yangyang Hu and Zhenxing Hu and Ke Huang and Ruiyuan Huang and Weixiao Huang and Zhiqi Huang and Tao Jiang and Zhejun Jiang and Xinyi Jin and Yu Jing and Guokun Lai and Aidi Li and C. Li and Cheng Li and Fang Li and Guanghe Li and Guanyu Li and Haitao Li and Haoyang Li and Jia Li and Jingwei Li and Junxiong Li and Lincan Li and Mo Li and Weihong Li and Wentao Li and Xinhang Li and Xinhao Li and Yang Li and Yanhao Li and Yiwei Li and Yuxiao Li and Zhaowei Li and Zheming Li and Weilong Liao and Jiawei Lin and Xiaohan Lin and Zhishan Lin and Zichao Lin and Cheng Liu and Chenyu Liu and Hongzhang Liu and Liang Liu and Shaowei Liu and Shudong Liu and Shuran Liu and Tianwei Liu and Tianyu Liu and Weizhou Liu and Xiangyan Liu and Yangyang Liu and Yanming Liu and Yibo Liu and Yuanxin Liu and Yue Liu and Zhengying Liu and Zhongnuo Liu and Enzhe Lu and Haoyu Lu and Zhiyuan Lu and Junyu Luo and Tongxu Luo and Yashuo Luo and Long Ma and Yingwei Ma and Shaoguang Mao and Yuan Mei and Xin Men and Fanqing Meng and Zhiyong Meng and Yibo Miao and Minqing Ni and Kun Ouyang and Siyuan Pan and Bo Pang and Yuchao Qian and Ruoyu Qin and Zeyu Qin and Jiezhong Qiu and Bowen Qu and Zeyu Shang and Youbo Shao and Tianxiao Shen and Zhennan Shen and Juanfeng Shi and Lidong Shi and Shengyuan Shi and Feifan Song and Pengwei Song and Tianhui Song and Xiaoxi Song and Hongjin Su and Jianlin Su and Zhaochen Su and Lin Sui and Jinsong Sun and Junyao Sun and Tongyu Sun and Flood Sung and Yunpeng Tai and Chuning Tang and Heyi Tang and Xiaojuan Tang and Zhengyang Tang and Jiawen Tao and Shiyuan Teng and Chaoran Tian and Pengfei Tian and Ao Wang and Bowen Wang and Chensi Wang and Chuang Wang and Congcong Wang and Dingkun Wang and Dinglu Wang and Dongliang Wang and Feng Wang and Hailong Wang and Haiming Wang and Hengzhi Wang and Huaqing Wang and Hui Wang and Jiahao Wang and Jinhong Wang and Jiuzheng Wang and Kaixin Wang and Linian Wang and Qibin Wang and Shengjie Wang and Shuyi Wang and Si Wang and Wei Wang and Xiaochen Wang and Xinyuan Wang and Yao Wang and Yejie Wang and Yipu Wang and Yiqin Wang and Yucheng Wang and Yuzhi Wang and Zhaoji Wang and Zhaowei Wang and Zhengtao Wang and Zhexu Wang and Zihan Wang and Zizhe Wang and Chu Wei and Ming Wei and Chuan Wen and Zichen Wen and Chengjie Wu and Haoning Wu and Junyan Wu and Rucong Wu and Wenhao Wu and Yuefeng Wu and Yuhao Wu and Yuxin Wu and Zijian Wu and Chenjun Xiao and Jin Xie and Xiaotong Xie and Yuchong Xie and Yifei Xin and Bowei Xing and Boyu Xu and Jianfan Xu and Jing Xu and Jinjing Xu and L. H. Xu and Lin Xu and Suting Xu and Weixin Xu and Xinbo Xu and Xinran Xu and Yangchuan Xu and Yichang Xu and Yuemeng Xu and Zelai Xu and Ziyao Xu and Junjie Yan and Yuzi Yan and Guangyao Yang and Hao Yang and Junwei Yang and Kai Yang and Ningyuan Yang and Ruihan Yang and Xiaofei Yang and Xinlong Yang and Ying Yang and Yi Yang and Yi Yang and Zhen Yang and Zhilin Yang and Zonghan Yang and Haotian Yao and Dan Ye and Wenjie Ye and Zhuorui Ye and Bohong Yin and Chengzhen Yu and Longhui Yu and Tao Yu and Tianxiang Yu and Enming Yuan and Mengjie Yuan and Xiaokun Yuan and Yang Yue and Weihao Zeng and Dunyuan Zha and Haobing Zhan and Dehao Zhang and Hao Zhang and Jin Zhang and Puqi Zhang and Qiao Zhang and Rui Zhang and Xiaobin Zhang and Y. Zhang and Yadong Zhang and Yangkun Zhang and Yichi Zhang and Yizhi Zhang and Yongting Zhang and Yu Zhang and Yushun Zhang and Yutao Zhang and Yutong Zhang and Zheng Zhang and Chenguang Zhao and Feifan Zhao and Jinxiang Zhao and Shuai Zhao and Xiangyu Zhao and Yikai Zhao and Zijia Zhao and Huabin Zheng and Ruihan Zheng and Shaojie Zheng and Tengyang Zheng and Junfeng Zhong and Longguang Zhong and Weiming Zhong and M. Zhou and Runjie Zhou and Xinyu Zhou and Zaida Zhou and Jinguo Zhu and Liya Zhu and Xinhao Zhu and Yuxuan Zhu and Zhen Zhu and Jingze Zhuang and Weiyu Zhuang and Ying Zou and Xinxing Zu},
      year={2026},
      eprint={2602.02276},
      archivePrefix={arXiv},
      primaryClass={cs.CL},
      url={https://arxiv.org/abs/2602.02276}, 
}

@misc{wasim2024videogroundingdinoopenvocabularyspatiotemporalvideo,
      title={Video-GroundingDINO: Towards Open-Vocabulary Spatio-Temporal Video Grounding}, 
      author={Syed Talal Wasim and Muzammal Naseer and Salman Khan and Ming-Hsuan Yang and Fahad Shahbaz Khan},
      year={2024},
      eprint={2401.00901},
      archivePrefix={arXiv},
      primaryClass={cs.CV},
      url={https://arxiv.org/abs/2401.00901}, 
}

@misc{jin2022embracingconsistencyonestageapproach,
      title={Embracing Consistency: A One-Stage Approach for Spatio-Temporal Video Grounding}, 
      author={Yang Jin and Yongzhi Li and Zehuan Yuan and Yadong Mu},
      year={2022},
      eprint={2209.13306},
      archivePrefix={arXiv},
      primaryClass={cs.CV},
      url={https://arxiv.org/abs/2209.13306}, 
}

@misc{lin2022stvgformerspatiotemporalvideogrounding,
      title={STVGFormer: Spatio-Temporal Video Grounding with Static-Dynamic Cross-Modal Understanding}, 
      author={Zihang Lin and Chaolei Tan and Jian-Fang Hu and Zhi Jin and Tiancai Ye and Wei-Shi Zheng},
      year={2022},
      eprint={2207.02756},
      archivePrefix={arXiv},
      primaryClass={cs.CV},
      url={https://arxiv.org/abs/2207.02756}, 
}

@misc{yang2025multiobjecttrackingretrievalllavavideo,
      title={Multi-Object Tracking Retrieval with LLaVA-Video: A Training-Free Solution to MOT25-StAG Challenge}, 
      author={Yi Yang and Yiming Xu and Timo Kaiser and Hao Cheng and Bodo Rosenhahn and Michael Ying Yang},
      year={2025},
      eprint={2511.03332},
      archivePrefix={arXiv},
      primaryClass={cs.CV},
      url={https://arxiv.org/abs/2511.03332}, 
}

@misc{yamaguchi2017spatiotemporalpersonretrievalnatural,
      title={Spatio-temporal Person Retrieval via Natural Language Queries}, 
      author={Masataka Yamaguchi and Kuniaki Saito and Yoshitaka Ushiku and Tatsuya Harada},
      year={2017},
      eprint={1704.07945},
      archivePrefix={arXiv},
      primaryClass={cs.CV},
      url={https://arxiv.org/abs/1704.07945}, 
}

@misc{vasudevan2018objectreferringvideoslanguage,
      title={Object Referring in Videos with Language and Human Gaze}, 
      author={Arun Balajee Vasudevan and Dengxin Dai and Luc Van Gool},
      year={2018},
      eprint={1801.01582},
      archivePrefix={arXiv},
      primaryClass={cs.CV},
      url={https://arxiv.org/abs/1801.01582}, 
}

@misc{gavrilyuk2018actoractionvideosegmentation,
      title={Actor and Action Video Segmentation from a Sentence}, 
      author={Kirill Gavrilyuk and Amir Ghodrati and Zhenyang Li and Cees G. M. Snoek},
      year={2018},
      eprint={1803.07485},
      archivePrefix={arXiv},
      primaryClass={cs.CV},
      url={https://arxiv.org/abs/1803.07485}, 
}

@misc{tan2022augmented2dtantwostageapproach,
      title={Augmented 2D-TAN: A Two-stage Approach for Human-centric Spatio-Temporal Video Grounding}, 
      author={Chaolei Tan and Zihang Lin and Jian-Fang Hu and Xiang Li and Wei-Shi Zheng},
      year={2022},
      eprint={2106.10634},
      archivePrefix={arXiv},
      primaryClass={cs.CV},
      url={https://arxiv.org/abs/2106.10634}, 
}

@misc{gu2024contextguidedspatiotemporalvideogrounding,
      title={Context-Guided Spatio-Temporal Video Grounding}, 
      author={Xin Gu and Heng Fan and Yan Huang and Tiejian Luo and Libo Zhang},
      year={2024},
      eprint={2401.01578},
      archivePrefix={arXiv},
      primaryClass={cs.CV},
      url={https://arxiv.org/abs/2401.01578}, 
}

\newpage
\appendix
\section{Prompt Templates and Reproducibility Materials}
\label{app:prompts}

This appendix provides the prompt templates used for evaluation, tool-style grounding calls, and Gemini-based SFT data synthesis. Reproduction materials, including code, configuration files, scripts, and instructions for accessing public benchmark data, will be released with the project page.

\subsection{Evaluation Prompt}
The evaluation prompt asks the model to perform second-level object tracking conditioned on a target description and to return a parseable JSON object with globally consistent labels.
\begin{verbatim}
Based on the description of the objects appearing in the video
"target to be tracked", please track the objects corresponding to this
description at every second (tracks per second) of the given video, and
provide the bounding box and a globally consistent label for each object.
Note that the same object across different timestamp should be assigned
the same label.

Note: Please organize the output in valid JSON format, where each key is
the time index (starting from 0), and the value is a list of detected
objects in current timestamp. Each element in the list corresponds to a
detection result in JSON format, with keys label and bbox_2d, whose values
are the object label and bounding box, respectively. For example:
{"0": [{"label": "car-1", "bbox_2d": [1,2,3,4]},
       {"label": "car-2", "bbox_2d": [2,3,4,5]}],
 "1": [{"label": "car-2", "bbox_2d": [4,5,6,7]},
       {"label": "person-1", "bbox_2d": [10,20,30,40]}]}
\end{verbatim}

\subsection{OpenClaw Case Prompt}
For the OpenClaw integration case, the model is queried with a shorter target-specific tracking prompt:
\begin{verbatim}
Output the object tracking results for target to be tracked in the video.
\end{verbatim}

\subsection{Gemini SFT Data Synthesis Prompt}
This prompt is used to synthesize chain-of-thought style decision traces for SFT. The generated trace is used for semantic reasoning supervision, while temporal spans and coordinates are replaced with ground-truth annotations before training.
\begin{quote}
\small
\textbf{Instruction.} Given the sampled video context, query, and available annotation metadata, write a concise reasoning trace for spatio-temporal grounding. Focus on identifying the referred target, discriminative attributes, relevant actions, distractors, and the rough temporal evidence.

\textbf{Important constraint.} Do not produce final coordinate values as authoritative labels. Coordinates and temporal boundaries will be replaced with ground-truth annotations in post-processing.

\textbf{Desired response.} Provide a time-then-space reasoning trace followed by a structured placeholder answer containing the target identity, approximate event interval, and trajectory fields.
\end{quote}

\subsection{Reproduction Materials}
The code and reproduction materials will be released with the project page. The materials contain environment setup, data preparation instructions, training/evaluation commands, prompt files, configuration files, and scripts for reproducing the main tables and figures.

\end{document}